\documentclass[letterpaper, 10 pt, conference]{ieeeconf}  
\IEEEoverridecommandlockouts                              
\usepackage[ruled,vlined]{algorithm2e}

\usepackage{subfigure}
\usepackage{booktabs}
\usepackage{balance}
\usepackage{color}
\usepackage{tabularx}
\usepackage{graphicx,dblfloatfix} 
\usepackage{amsmath}
\usepackage{amssymb}
\usepackage{listings}
\usepackage{textcomp}
\usepackage{url}
\usepackage{multirow}
\usepackage{todonotes}
\usepackage{hyperref}
\usepackage{cleveref}
\usepackage{wrapfig}
\crefformat{footnote}{#2\footnotemark[#1]#3}
\usepackage{ragged2e} 
\usepackage[subnum]{cases}
\newcolumntype{Y}{>{\RaggedRight\arraybackslash}X} 
\usepackage[noend]{algpseudocode}

\usepackage{tikz}
\usepackage{tkz-euclide}
\usetikzlibrary{math}
\usepackage{pgfplots}
\usetikzlibrary{arrows}

\usepackage{float}

\usepackage{multicol, blindtext}

\def\fps@figure{htp}
\def\fps@table{htp}

\newcommand{\bi}{\begin{itemize}}
\newcommand{\ei}{\end{itemize}}

\newcommand{\bfig}{\begin{figure}}
\newcommand{\efig}{\end{figure}}

\newcommand{\benum}{\begin{enumerate}}
\newcommand{\eenum}{\end{enumerate}}

\newcommand{\be}{\begin{equation}}
\newcommand{\ee}{\end{equation}}

\newcommand{\ba}{\begin{eqnarray}}
\newcommand{\ea}{\end{eqnarray}}

\title{\LARGE \bf
   Virtual Surfaces and Attitude Aware Planning and Behaviours for Negative Obstacle Navigation*\thanks{* This research was, in part funded by the US Government under the DARPA Subterranean Challenge. The views, opinions, and findings expressed are those of the authors and should not be interpreted as representing the official views or policies of the Department of Defense or the U.S. Government. Approved for Public Release, Distribution Unlimited.}
}

\author{Thomas Hines$^{1}$, Kazys Stepanas$^{1}$, Fletcher Talbot$^{1}$, Inkyu Sa$^{1}$, Jake Lewis$^{2}$\\ Emili Hernandez$^{3}$, Navinda Kottege$^{1}$, and Nicolas Hudson$^{1}$  
\thanks{$^{1}$ T. Hines, K. Stepanas, F. Talbot, I. Sa, N. Kottege and N. Hudson are with the Robotics and Autonomous Systems
Group (RASG), CSIRO, Pullenvale, QLD 4069, Australia. All correspondence should be addressed to {\tt\small inkyu.sa@csiro.au}}
\thanks{$^{2}$ J. Lewis is with Euclideon Holographics, Murarrie QLD 4172, Australia and was a research intern at the RASG at the time of this work.}
\thanks{$^{3}$ E. Hernandez is with Emesent, Milton, QLD 4064, Australia and was with the RASG at the time of this work.}%
}

\begin{document}
\maketitle
\markboth{IEEE Robotics and Automation Letters. Submitted Version. Pending review}
{TBD \MakeLowercase{\textit{et al.}}: Shared Virtual Surfaces for Negative Obstacle Navigation
} 

\begin{abstract}
This paper presents an autonomous navigation system for ground robots traversing aggressive unstructured terrain through a cohesive arrangement of mapping, deliberative planning and reactive behaviour modules. All systems are aware of terrain slope, visibility and vehicle orientation, enabling robots to recognize, plan and react around unobserved areas and overcome negative obstacles, slopes, steps, overhangs and narrow passageways. This is one of pioneer works to explicitly and simultaneously couple mapping, planning and reactive components in dealing with negative obstacles. The system was deployed on three heterogeneous ground robots for the DARPA Subterranean Challenge, and we present results in Urban and Cave environments, along with simulated scenarios, that demonstrate this approach. 
\end{abstract}


\section{Introduction}
\label{sec:introduction}


Negative obstacles such as cliffs, ditches, depressions, pose a difficult problem for autonomous navigation in unstructured environments, as is reported from deployments of highly capable vehicles in natural terrain, such as Crusher \cite{stentz2007crusher} or the Legged Squad Support System (LS3) \cite{bajracharya2013stereo}. Negative obstacles are difficult to detect from vehicle mounted sensors as the near-field terrain occludes a drop, slope and/or trailing rising edge. Compared to a positive obstacles (e.g., walls, people, boxes), occlusions and viewing angles result in fewer pixels-on-target \cite{Matthies2003Negative}, which in turn reduces the effective detection range, often to within the stopping distance of ground vehicles moving at any appreciable speed. 

For vehicles capable of autonomously traversing extreme ramp angles $>45^{\circ}$, even slowly approaching a negative obstacle requires significant care in planning and sensor placement to ensure safe navigation. The obstacle must be approached from an optimal angle, so the system can observe and determine if the terrain is a traversable ramp, or a fatal discontinuity. There are few examples of robotic systems capable of detecting and traversing negative obstacles in unstructured terrain, such as \cite{Sinha2013}, where the system can both detect and safely traverse gaps by reasoning about the boundaries of the gap or unobserved region.

\begin{figure}
\begin{center}
\includegraphics[width=\columnwidth]{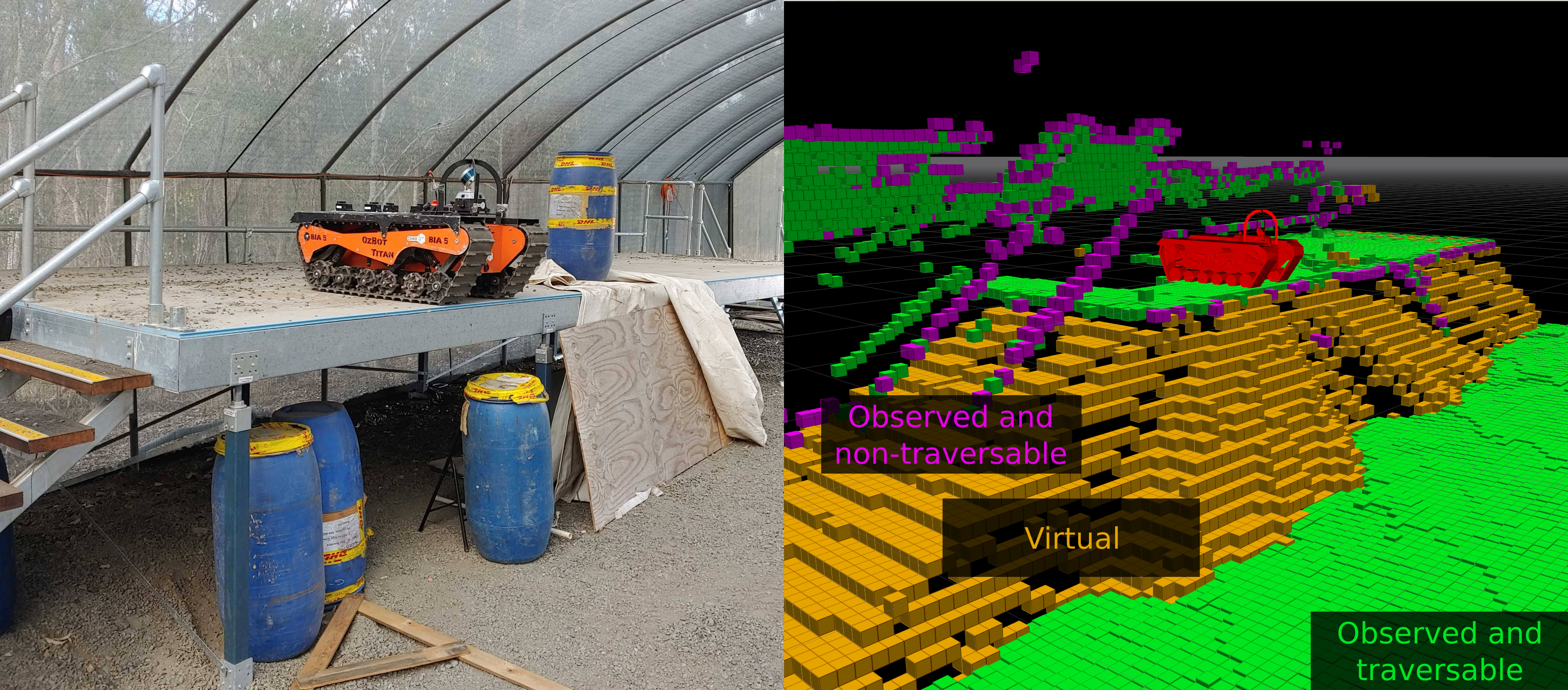}
\end{center}
\vspace{-5mm}
\caption{A robot on the edge of a negative obstacle (left) and the corresponding costmap (right). Cells are coloured by type: green indicates an observed and possibly traversable surface; magenta is observed and definitely non-traversable (fatal); orange is virtual.
}
\label{fig:experiments-real-fatal-photo}
\vspace{-5mm}
\end{figure}

We adopt a \emph{virtual surface} concept to estimate the best case slope \emph{within} occluded regions of the map as shown in \Cref{fig:experiments-real-fatal-photo}. These virtual surfaces are used to plan trajectories. By updating virtual surfaces in real-time, while continuously planning and collecting data, negative obstacles can be safely approached and avoided if they are found to be unsafe. The methods of planning and control will be summarised but the details and analysis of these methods are not the subject of this paper and will not be addressed. The primary contributions of this paper are:

\begin{enumerate}
  \item An open-source 3D probabilistic voxel occupancy map \cite{ohm} that uses ray tracing for virtual surface construction, suitable for real-time robot navigation, and relevant data sets for evaluation
  \footnote{\url{https://doi.org/10.25919/32q7-9e58}}.


  \item Relevant field trial results from urban and cave environments, demonstrating the ability to navigate negative obstacles in extreme terrain. 

\end{enumerate}

We present literature reviews for relevant prior work in \Cref{sec:related work}. \Cref{sec:method} addresses methodologies proposed in this paper including the sensor pack, mapping and heightmap generation, local costmap generation, path planning, and velocity generation. \Cref{sec:experiments} contains experimental results in both real-world subterranean and simulated environments. Lessons learned from field experiments are shared and discussed in \Cref{sec:discussions}. Lastly, \Cref{sec:conclusions} summarises feasible outlooks of the proposed work.

\section{Related work}
\label{sec:related work}
In this section we summarise some relevant prior work in occupancy map-based heightmap representation, negative obstacle modelling, handling occlusions, and unmanned ground vehicle (UGV) path planning.

When handling negative and positive obstacles it is useful to have an effective heightmap representation. This encapsulates each grid cell’s height to be used for estimating terrain properties.

Octomap\footnote{http://wiki.ros.org/octomap} \cite{hornung13auro} is a popular library that provides a 3D probabilistic occupancy map using an octree representation for memory efficiency. Octomap has had numerous successful deployments including underwater \cite{Sodhi2019-bo}, flying \cite{Pestana2019-la}, walking \cite{Fankhauser2018-fa} and ground \cite{Wang2017-ed} applications.

Drawing inspiration from \cite{hornung13auro}, we implemented an efficient heightmap generation algorithm \cite{ohm} extracted from a 3D voxel occupancy map computed in GPU bound to a predefined local area. Technical detail is presented in \Cref{sec:OHM}.




Positive obstacles can be effectively detected using the eigenvectors of the covariance matrix of the (x, y, z) coordinates of cells in a small patch of the heightmap \cite{Rusu2011-gp}. In contrast, negative obstacles are often unobservable and are inferred from gaps in the map, thus detection using geometric information is prevalent. Often depth data is accumulated into a 3D occupancy map and unobserved cells are classified as obstacles based on adjacent observed cells. 

Camera based methods of detecting negative obstacles have been demonstrated. \cite{coughlan2007terrain} used depth from stereo vision to find drop-offs but did not attempt to bound and remember slope in occluded regions. \cite{bajracharya2013stereo} used an heightmap from stereo vision and traced rays originating at the camera along the ground to bound the slope and step height in occluded regions. This is the most similar existing method to ours. \cite{Ghani2017-uy} used a pointcloud from a RGBD camera with a similar method by finding a floor plane and interpolating the slope between points below the plane and points on the plane that represent the edge of the negative obstacle. \cite{murarka2008detecting} used stereo vision and motion cues to extract planes from observed surfaces or find the difference between features on either side of edges in a depth image. These methods can be susceptible to lighting changes and either assume the world consists of planes or work along rays between the sensor and the obstacle while our approach does not have these limitations. Additionally, our work integrates the rays from the sensor to the return surfaces implicitly and maintains this information even as the robot moves away to allow for negative obstacle slope bounds to be remembered.

Thermal imagery has been exploited by observing that negative depressions remain warmer than surrounding terrain at night \cite{Matthies2003Negative}. This thermal intensity data was later used in combination with stereo vision and geometry based thresholds to further increase detection range \cite{Rankin2007}.

Lidar sensors have been used \cite{Shang2014-ev,Shang2016-yq} with 3D ray tracing to determine occlusion and classify obstacles using nearby observed voxels \cite{heckman2007} or classify the points below the ray path \cite{larson2011} with heuristics and SVM. Our work instead estimates a 3D virtual surface inside the occlusion, as opposed to using just the edges of the occlusion. In addition our method of determining traversability considers the vehicle footprint and is orientation dependent.

Lidar based detection is also achieved in heightmaps \cite{morton2011positive} by propagating information from all nearby observed cells to infer the unobserved terrain. 

Range sensors have been used to avoid negative obstacles in \cite{devigne2019shared,simpson2004smart}. Our work uses a lidar capable of making higher detail observations of the environment than the systems presented in both of these papers. Our work also maintains an occupancy map which provides a memory of recently discovered negative obstacles to assist with higher degrees of autonomy capable of exploring drop offs in order to discern between occluded traversable slopes and non-traversable negative obstacles.

\cite{zhang2020method,darbari2017dynamic} present methods of planning paths to fully observe 3D spaces accounting for occlusion by obstacles. These are both targeted at UAVs but could be adapted to explore and discover occluded slopes and negative obstacles. However, they do not directly address the prediction of UGV traversability of occluded slopes which is a key feature of our work.

Planning through regions occluded by positive obstacles has been studied such as in \cite{yu2019occlusion}. This work predicted positive obstacles in flat environments and was not intended to address negative obstacles.

Learning-based methods of scene prediction such as in \cite{wang2020learning} can be used to predict occupancy in obscured regions. Traversability of these obscured regions could therefore also be predicted. \cite{wang2020learning} trained for accuracy not optimism which could encourage false positive predictions of fatal negative obstacles. Our work makes explicitly optimistic predictions of the occluded surfaces which encourages autonomous exploration of edges. Learning-based methods can be trained for optimism in which case they could be usable alternatives to some of our work.

Finally, a motion planning algorithm uses the heightmap and costmap (including positive and negative obstacles) to compute a trajectory to safely reach a goal. Motion planning is an important building block for autonomous vehicles so there has been active research on this topic for decades \cite{Ferguson2008-tt}.

Motion planning often takes into account not only geometric details of paths (e.g. Dijkstra \cite{Dijkstra1959-jz} or $A^*$ \cite{Harel1987-cl}) but also the vehicle’s kinematic constraints in order to generate feasible trajectories \cite{Dolgov2010}. In this paper we used an hybrid $A^*$ \cite{Dolgov2010} planner which is a variant of the $A^*$ algorithm that considers the vehicle's kinematic constraints. The planner computes a desired trajectory using a simplified vehicle configuration evaluation method (see \cite{Otsu2020-tm} for a more capable method) to ensure the vehicle is capable of traversing the terrain below it. The planner is used to plan onto virtual surfaces at a range, and has an appropriate sensor configuration such that fatality of negative obstacles can be determined if approached head on before traversal.

\section{Methodologies}
\label{sec:method}

\begin{figure}
\begin{center}
\includegraphics[width=0.7\columnwidth]{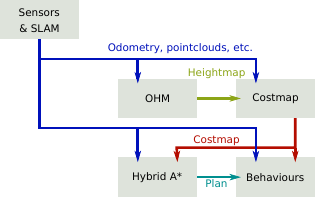}
\end{center}
\vspace{-5mm}
\caption{A flow diagram including the components mentioned in this paper and the information shared between them.}
\label{fig:vs-flow}
\vspace{-5mm}
\end{figure}

This section contains descriptions of how the key components contribute to handling negative obstacles using virtual surfaces. \Cref{fig:vs-flow} shows which of these components generate and use the relevant pieces of information.

\subsection{Sensing and SLAM payload}
\label{sec:pack}
\begin{figure}[!b]
\begin{center}
\vspace{-0.5cm}
\includegraphics[width=\columnwidth]{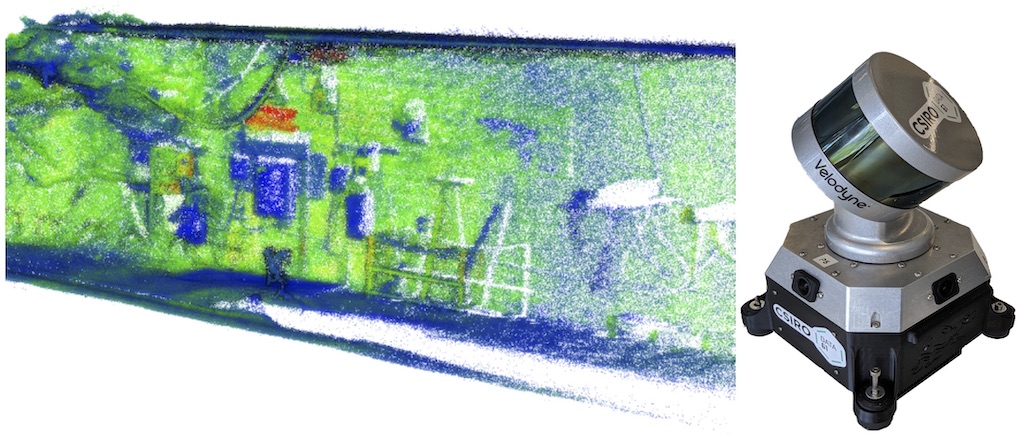}
\end{center}
\caption{
A sample point cloud produced by a Catpack sensor suite (left) and the sensing payload (right).
}
\label{fig:catpack}
\end{figure}

The sensing payload used in all experiments described here (the ``Catpack" in \Cref{fig:catpack}) consists of a tilted Velodyne VLP-16 lidar on a rotating, encoder tracked mount, a Microstrain CV5-25 IMU and a custom timing board used for time synchronisation between sensors. This 0.5\,Hz rotating lidar mount improves sensor coverage around the vehicle, while the tilt angle improves visibility of the ground in front of the vehicle and the lidar coverage density and diversity. The payload runs custom SLAM software \cite{Bosse2012-ku}. The Catpack is placed near the front of the vehicle ensuring a downward field of view greater than the maximum ramp angle the vehicle can traverse.

The Catpack publishes odometry and a raw lidar pointcloud via ROS \cite{ros}. The pack also localises the pointcloud to account for the encoder rotation and lidar orientation. As a consequence, these points are published in the vehicle frame at ~290k points/s, depending on the environment, while the local odometry pose is updated at 100Hz with higher accuracy poses generated in the data stream at approximately 4Hz.

\subsection{Mapping and heightmap generation}
\label{sec:OHM}


The heightmap is extracted from a 3D probabilistic voxel occupancy map generated by the Occupancy Homogeneous (OHM \cite{ohm}) library. This library was used for its ability to process high rate lidar data with GPU operations. The occupancy map is generated from the pointcloud and local odometry pose output of the Catpack. Each ray is integrated into the occupancy map using a technique which supports normal distribution transforms such as in \cite{doi:10.1177/0278364913499415}. Each voxel is classified as being occupied, free or (by default) unobserved. The map is generated at a 10cm resolution.

Since the occupancy map is generated from local SLAM odometry, the map is vulnerable to global misalignment errors. That is to say, this map does not consider any global optimisations such as those enacted by loop closure algorithms. This is addressed by only maintaining the occupancy map locally around the vehicle ($\sim$10m$\times$10m) where misalignment errors are insignificant.

We assume the map is vertically aligned with the $z$-axis when generating the heightmap and examine individual voxel columns within the occupancy map. It is during heightmap generation that we add an additional voxel classification, virtual surface voxels, which occur at the interface between free and unobserved space. Specifically, a virtual surface classification is assigned to free voxels which have an unobserved voxel immediately below. A virtual surface represents a best case surface in regions which are shadowed in the map and cannot be adequately observed. Virtual surface voxels are only used when there are no occupied voxel candidates within the search constraints of a column.

\begin{figure}
    \centering
    \includegraphics[width=45mm]{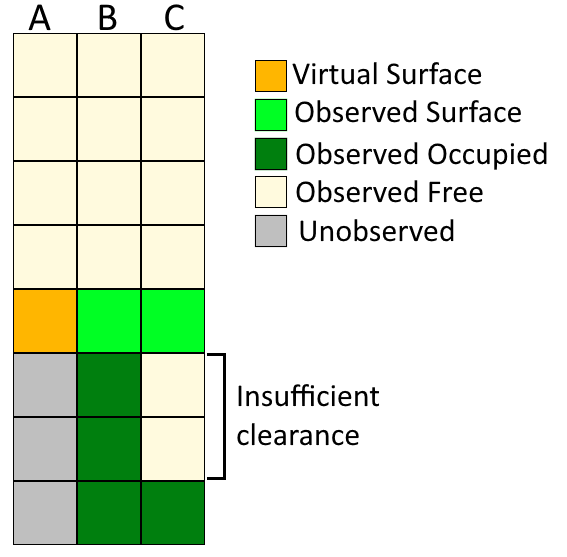}
    \vspace{-4mm}
    \caption{Example columns considered for heightmap generation shown from a side on perspective. Column A depicts a virtual surface generated at the interface between known free and unobserved voxels. Columns B and C show known surfaces, where column C also shows a surface candidate rejected due to insufficient clearance.}
    \label{fig:heightmap-columns}
    \vspace{-6mm}
\end{figure}

In addition to the voxel classification, we also impose a clearance constraint to ensure there is a vertical space large enough for the vehicle to fit through. For this constraint we ensure there are sufficient free or unobserved voxels above each candidate voxel to meet the clearance constraint. Three columns considered for heightmap generation are shown in \Cref{fig:heightmap-columns}.

\begin{figure}[t]
\begin{center}
\includegraphics[width=0.8\columnwidth]{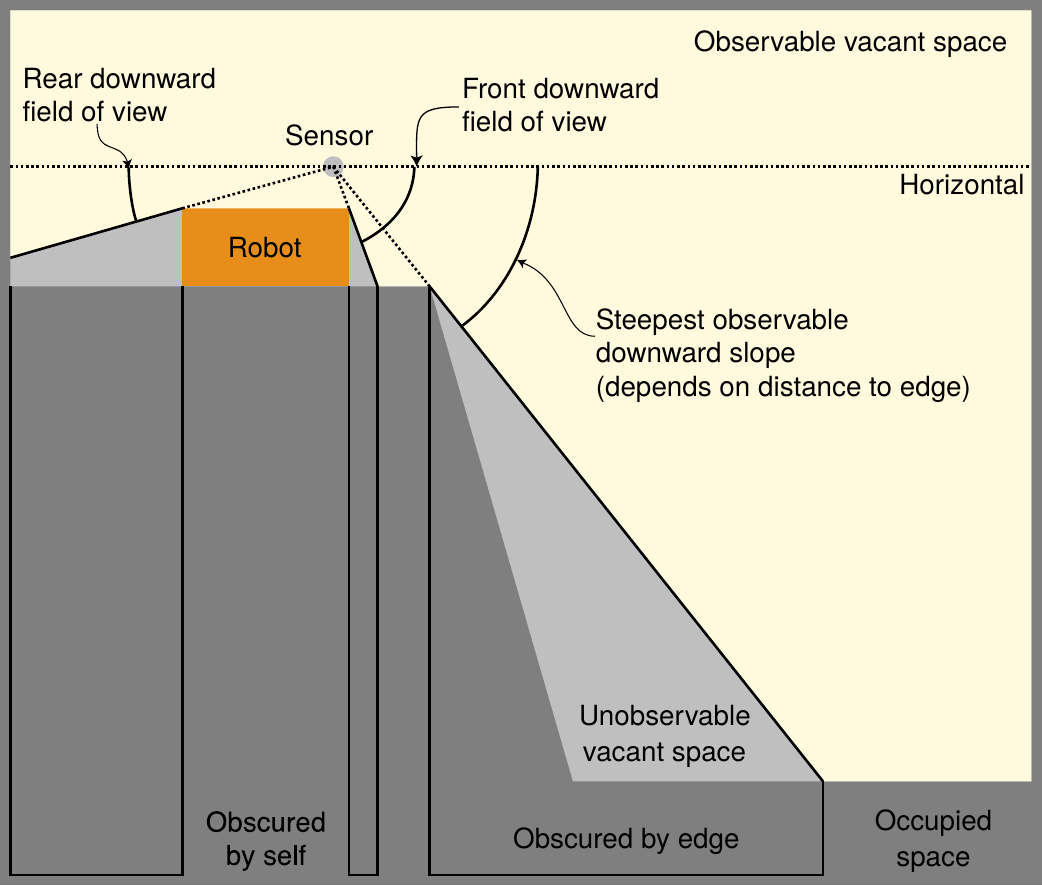}
\end{center}
\vspace{-5mm}
\caption{
Side view of a robot near the edge of a negative obstacle. The field of view is limited due to obstruction by the robot's own body and by the edge of negative obstacles.
}
\label{fig:virtual-surface-field-of-view}
\vspace{-3mm}
\end{figure}

Virtual surfaces are a key input to detecting negative obstacles. A virtual surface is any region of voxels in the heightmap consisting of virtual voxels as described above. A virtual surface represents a region of observational uncertainly and the best case slope for that region. Such surfaces will often come about from shadowing effects in sensor observations, but will also occur when real surface observations cannot be made such as around black bodies and water. Various scenarios can be seen in \Cref{fig:virtual-surface-field-of-view} where virtual surfaces are generated for a downward slope. As the vehicle approaches the edge of the slope, real observations may be made and a real surface can be generated in the heightmap.

\Cref{fig:virtual-surface-field-of-view} also shows two other factors that impact observations of negative obstacles:

\begin{itemize}

    \item the downward field of view is much better in front of the robot because the sensor is closer to the front edge of its body;

    \item the maximum steepness of the downward slope that a robot is able to observe depends on the distance between the sensor and the edge of the slope as well as the height of the sensor.

\end{itemize}

\begin{figure}[t]
\begin{center}
\includegraphics[width=0.7\columnwidth]{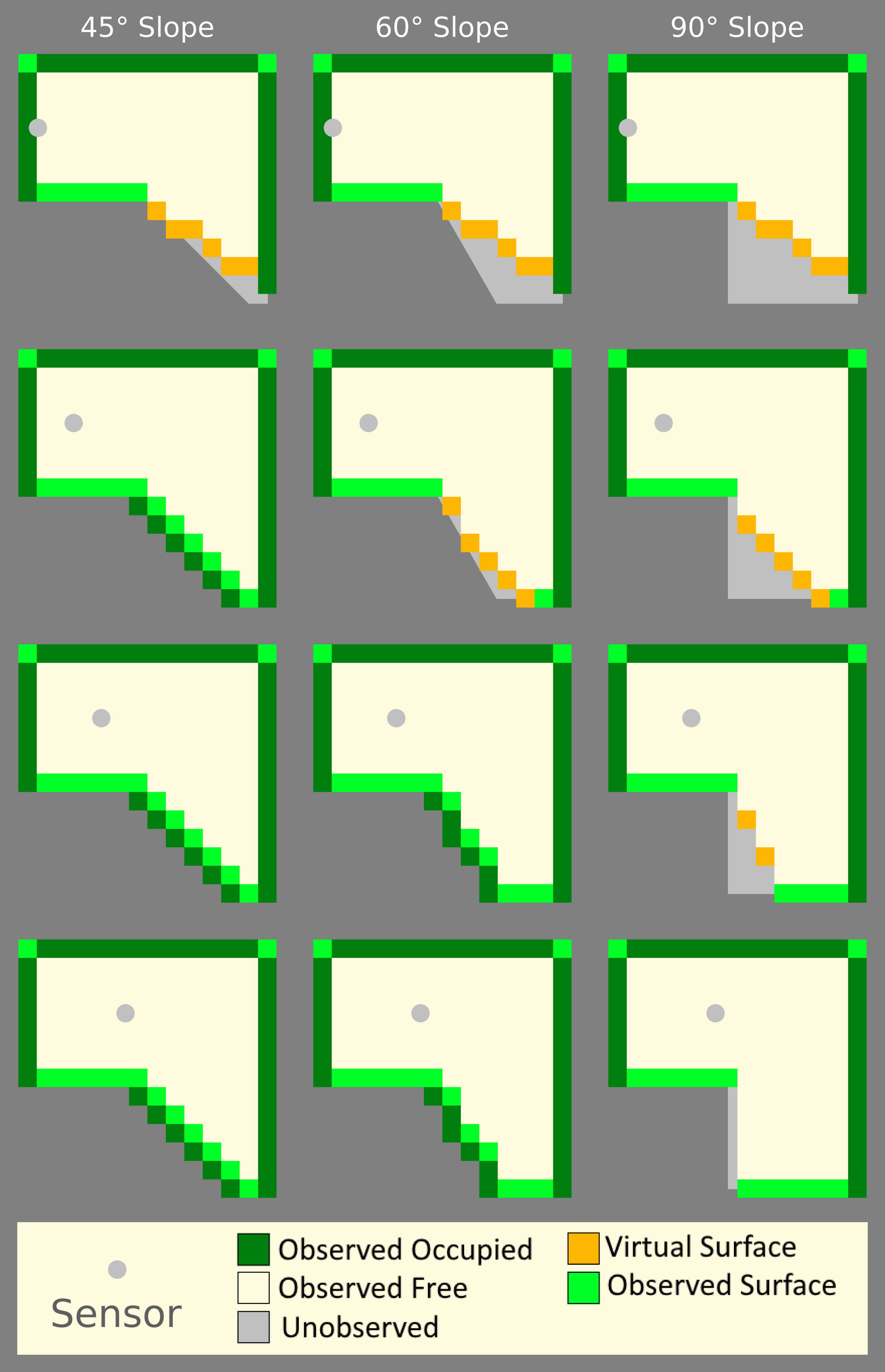}
\end{center}
\vspace{-5mm}
\caption{Side views of a robot approaching the edges of negative obstacles and the heightmaps that might be generated in each case. Each column shows a slope of different steepness. Each row shows the robot at different distances to the edge.
}
\label{fig:virtual-surface-approach}
\vspace{-5mm}
\end{figure}

\Cref{fig:virtual-surface-field-of-view,fig:virtual-surface-approach} highlight the uncertainty inherent with using virtual surfaces: they are by definition an upper bound for the surface beneath them. In \Cref{fig:virtual-surface-approach}, a virtual surface is initially generated when the vehicle is too far away from the edge to be able to observe the slope beyond it. As the vehicle approaches the edge, the slope of the virtual surface increases until the real slope is directly observed. There is a limit to the downward field of view, as described in \Cref{fig:virtual-surface-field-of-view}. If the slope is steeper than the best downward field of view then the virtual surface will continue to increase in steepness until it reaches that limit and no real observation of the slope will ever be made. The conclusion drawn is that a virtual surface cannot be confirmed as a negative obstacle until the vehicle is in close proximity to the potential surface and has sufficient downward field of view.





\subsection{Costmap generation}
\label{sec:costmap}


OHM produces a 2.5D heightmap that labels each cell as either real (observed), virtual (best case inferred) or unknown. Within this heightmap it is possible to identify obstacles that cannot be traversed in any possible way (e.g. the walls of a hallway). These obstacles are identified and labelled in the costmap. The costmap contains the same information as the heightmap but with additional labels for non-fatal (possibly traversable) and fatal (definitely non-traversable) cells. \Cref{fig:experiments-real-fatal-photo} shows a visualisation of the costmap: the handrail of the stairs and the edges of the platform are labelled fatal; the ground and the flat surface of the platform are labelled non-fatal; and the shadow of the platform is labelled virtual.

Virtual surfaces are treated mostly as if they were real, observed surfaces. This is because they represent the best case slope (shallowest possible) they could contain. If the best case slope is traversable, then they can only be labelled as non-fatal (possibly traversable). If the best case slope is non-traversable then they should be labelled as fatal (definitely non-traversable).

We used the GridMap \cite{Fankhauser2016GridMapLibrary} library to arrange a series of filters to generate the costmap. The filter used to identify fatal obstacles was similar to the one used in \cite{Zhao2020-sz} with an added step to remove small concavities that the vehicle can easily traverse and to use vertical sections in three directions instead of four.

\subsection{Hybrid $A^*$ and the dynamic approach of negative obstacles}
\label{sec:astar}







An implementation of hybrid A* ~\cite{Dolgov2010} was used to plan robot trajectories to move toward goals while avoiding obstacles ~\cite{Nemec2019}. The cost function was modified to consider heightmap and costmap cells that intersect with the robot footprint at each configuration. The heightmap cells include virtual surfaces and are used to estimate the pitch and roll components of the robot's orientation at each configuration so that terrain steepness can be accounted for when planning. As virtual surfaces are the best case slope for unobserved regions, plans can be made to use virtual surfaces until they are too steep to traverse.

From a distance, a virtual surface cannot be identified as traversable or non-traversable. By planning as if distant virtual surfaces are traversable, the robot will approach virtual surfaces that could provide a helpful route (e.g., ramps or stairs). During the approach, if the virtual surface contains a fatal negative obstacle (e.g., a pit or cliff), the slope of the virtual surface will become fatally steep. In this situation the cost function should be sufficient to avoid configurations on the virtual surface. Hybrid A* will only generate routes to the goal that avoid the negative obstacle. OHM is configured to maintain the local occupancy map indefinitely within a radius of about 5m. So long as the distance between the robot and a negative obstacle stays below this radius, the best observations of it will be remembered as the robot moves away.

\subsection{Velocity command generating behaviours}
\label{sec:behaviours}

A set of independent behaviours were used to generate velocity commands for the robot to execute. These behaviours run in parallel and each is designed to be activated for a specific task or in a specific situation. Each behaviour constantly monitors the current situation to check if it is admissible (i.e., can make progress on its task or can handle the current situation). Additionally, each admissible behaviour generates a velocity command for the robot. The behaviours are ordered by priority and the velocity commands generated by the highest priority admissible behaviour are executed.

Most of the robot's movement is performed by the \emph{path follow} behaviour. \emph{Path follow} attempts to follow the path generated by hybrid $A^*$. However, hybrid $A^*$ is not guaranteed to produce a path that can be safely followed. The robot itself is also not guaranteed to follow the hybrid $A^*$ path exactly. During testing, the robot may attempt unsafe manoeuvres, such as climbing excessively steep slopes. The \emph{path follow} behaviour is only admissible when the robot's pitch and roll are within some acceptable range. When the pitch or roll become excessive, the \emph{path follow} behaviour relinquishes control. The \emph{orientation correction} behaviour is admissible when the pitch or roll are excessive. \emph{Orientation correction} will take over when \emph{path follow} causes an extreme pitch or roll in order to prevent the robot from tipping over. Situations have also been encountered where hybrid A* cannot generate a plan, such as when there is fatal cost within the footprint of its first search node. In this case, \emph{path follow} will not be admissible because it has no path to follow. The \emph{decollide} behaviour will take control in order move the robot to a nearby position where hybrid A* can generate a plan.

\section{Experiments and results}
\label{sec:experiments}

\begin{figure}
\begin{center}
\includegraphics[width=0.9\columnwidth]{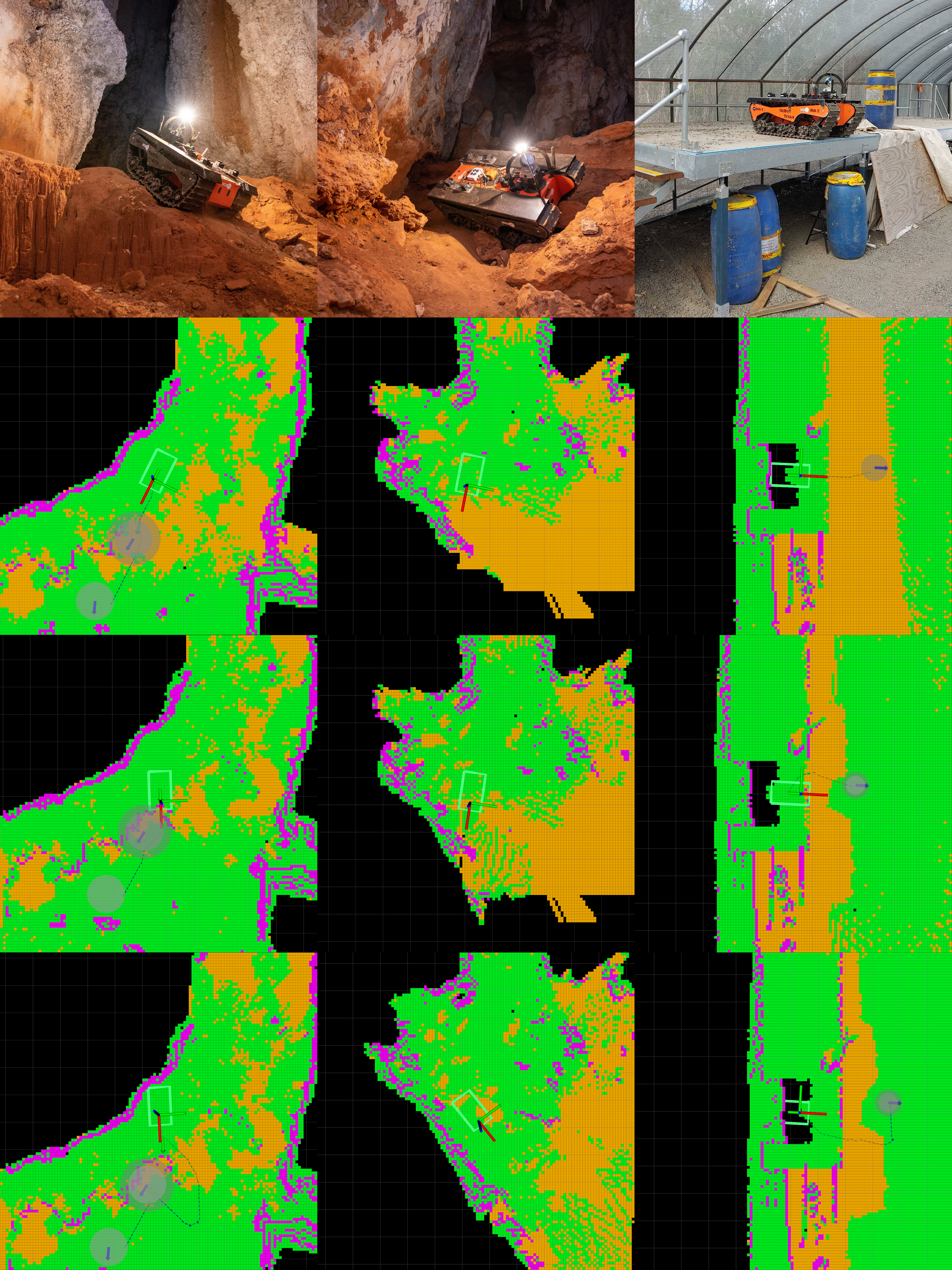}
\end{center}
\vspace{-5mm}
\caption{A real robot approaching various virtual surfaces. Each column is a different scenario. Each row shows the robot in a different position in chronological order. This figure is best viewed in colour.}
\label{fig:experiments-real}
\vspace{-7mm}
\end{figure}

\Cref{fig:experiments-real-fatal-photo,fig:experiments-real} contain photographs of the robot used to test the methodology described in this paper. The system was used in the DARPA SubT Urban event, some parts of the Chillagoe-Mungana Caves, test facilities at CSIRO's QCAT site and various other locations.

\Cref{fig:experiments-real} shows sample costmaps as three different virtual surfaces were approached. The photos were taken afterward and do not represent the actual position of the robot in any of the costmaps. The first two columns were recorded in the Chillagoe-Mungana Caves. The third column was recorded at CSIRO's QCAT site.

The first column depicts the robot approach the edge of a rock cliff with a traversable slope next to it. The goal was set to a position beyond the bottom of the cliff. Initially the cliff is represented by a traversable virtual surface. Hybrid A* generated a path to drive off the cliff. As it approached the virtual surface became steeper until it was non-traversable. After this, the robot backed up and hybrid A* generated a path to avoid the steepest part of the cliff.

The second column shows the robot approach a traversable downward slope. The original data that included the goal and hybrid A* paths was corrupted. The costmap was reproduced later based on raw data recorded from the sensors but the goal and paths were not. Initially the downward slope was virtual. During the approach the slope was observed and labelled as non-fatal.


The third column portrays the robot approaching the sharp edge of a platform. Again, hybrid A* initially planned off the edge. After getting close to the edge it was identified as non-traversable and the robot backed up and switched to a path that avoided it.

\subsection{Performance evaluation}
\label{sec:experiments-evaluation}

A robot may become damaged during a fall if it fails to avoid a negative obstacle. A Gazebo based simulation was used so that failures were acceptable. The simulation included analogues for most of the components in the system including models of the vehicle, Catpack and some environments. The vehicle model was geometrically accurate but made no attempt to emulate any dynamic mechanical components. The Catpack model had a tilted and rotating Velodyne that closely emulated a real Catpack. The accuracy of the geometry and sensor models made this simulation a useful tool during development and testing. The most significant inaccuracy of the simulation was in the environmental models. These were polygonal and did not emulate the natural complexity and roughness of real environments.

\begin{table}[t]
\centering
\caption{Success rates and execution durations for the simulated scenarios and methods show in \Cref{fig:experiments-simulation-fatal-trench-comparison,fig:experiments-simulation-ramp-comparison}}
\begin{tabularx}{\columnwidth}{ccrrr}
\toprule
\multirow{2}{*}{Method} & \multirow{2}{*}{Scenario} & \multirow{2}{*}{Success rate} & \multicolumn{2}{c}{Duration (s)} \\
& & & mean & std dev \\
\midrule
\multirow{2}{*}{Virtual surface} & Trench & 100\% & 60.9 & 0.0 \\
                                 &   Ramp & 100\% & 27.1 & 4.2 \\
\midrule
\multirow{2}{*}{Non-traversable} & Trench & 100\% & 60.9 & 0.0 \\
                                 &   Ramp &   0\% & 60.9 & 0.0 \\
\midrule
\multirow{2}{*}{    Traversable} & Trench &   0\% &  5.2 & 0.6 \\
                                 &   Ramp & 100\% & 25.4 & 3.7 \\
\bottomrule
\end{tabularx}
\label{table:experiments-simulation-values}
\vspace{-5mm}
\end{table}

\begin{figure}
\begin{center}
\includegraphics[width=\columnwidth]{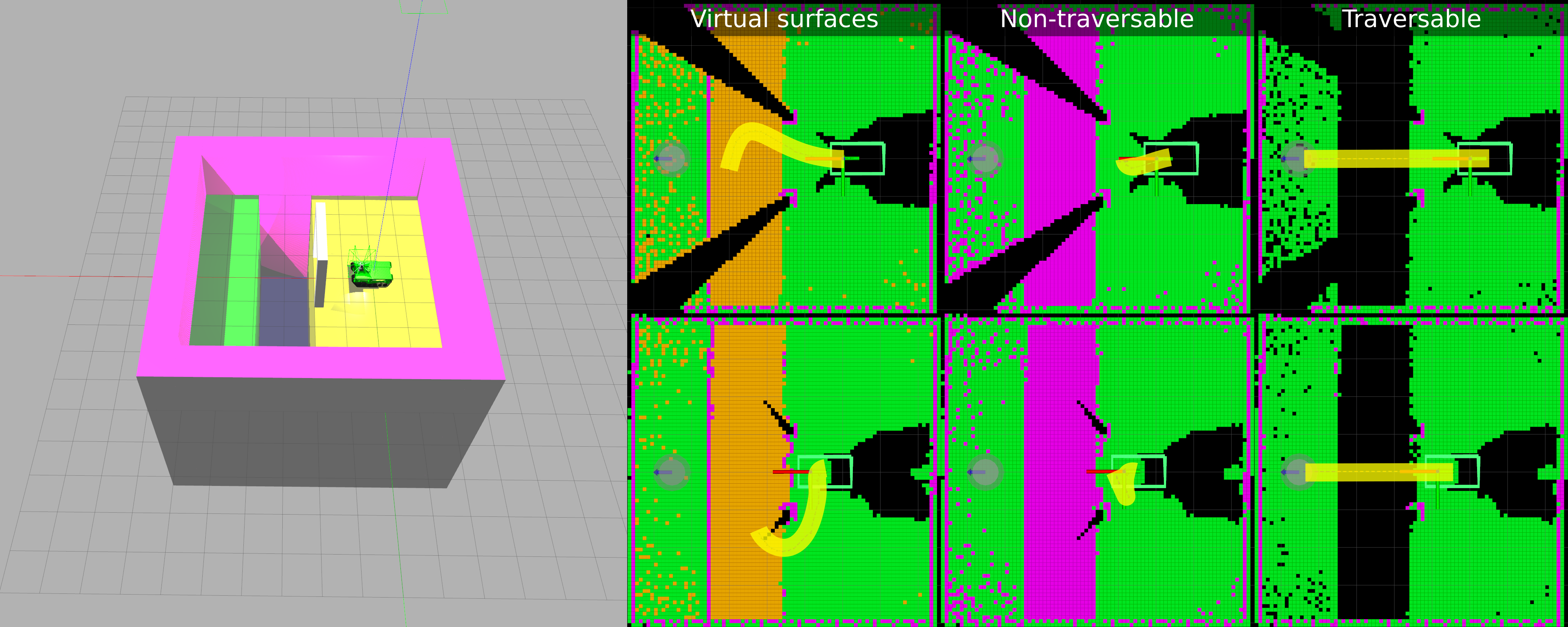}
\end{center}
\vspace{-5mm}
\caption{
A simulated robot approaching a fatal trench.
}
\label{fig:experiments-simulation-fatal-trench-comparison}
\vspace{-3mm}
\end{figure}

\begin{figure}
\begin{center}
\includegraphics[width=\columnwidth]{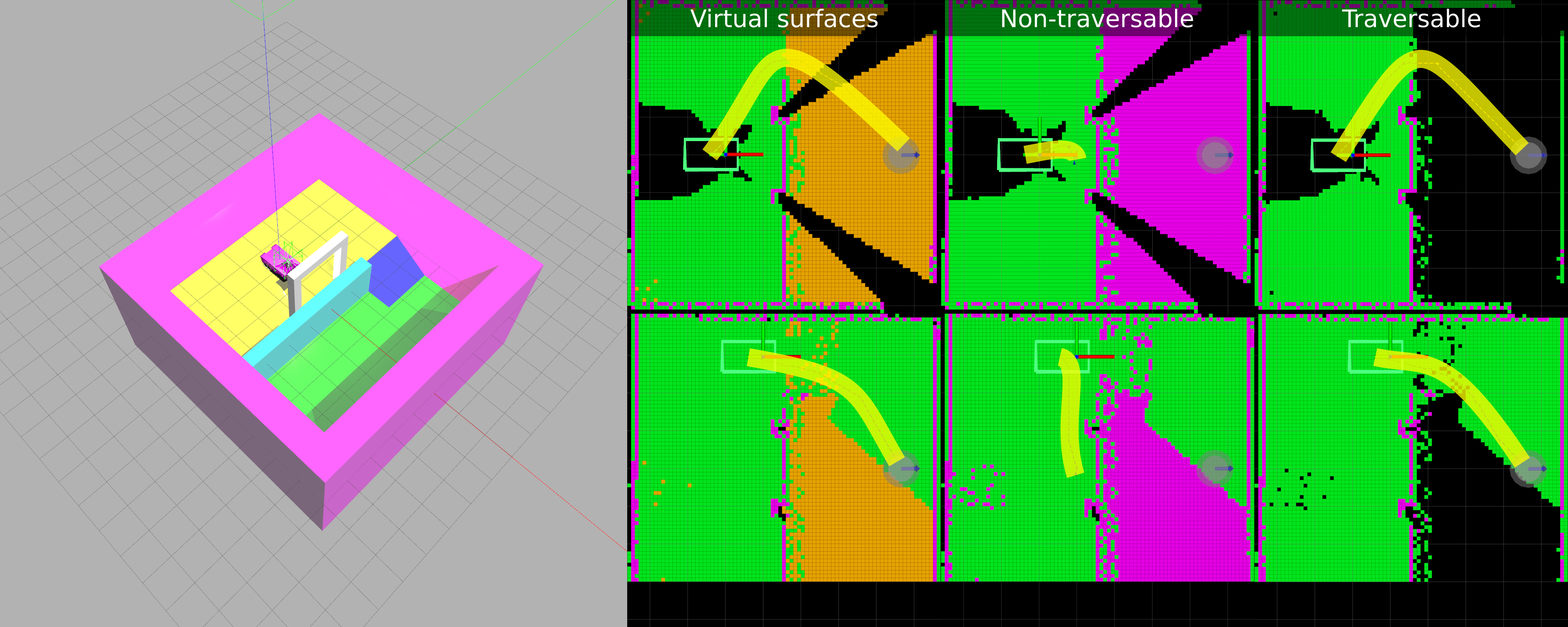}
\end{center}
\vspace{-5mm}
\caption{
A simulated robot that must drive down a ramp to reach the goal.
}
\label{fig:experiments-simulation-ramp-comparison}
\vspace{-5mm}
\end{figure}

\Cref{fig:experiments-simulation-fatal-trench-comparison,fig:experiments-simulation-ramp-comparison} contain images of simulated tests of two alternative ways of handling virtual surfaces: treating them as non-traversable and treating them as traversable. On the left of both figures is a render of the scenario. On the right are 3 columns of costmaps generated by handling virtual surfaces in the different ways. In each costmap: the green rectangle outline is the robot's current position; the grey filled circle is the goal; and the yellow line is the path generated by hybrid A*. These figures are best viewed in colour.

When virtual surfaces were considered non-traversable the planner never attempted to move toward the obscured downward ramp. The second column of costmaps in \Cref{fig:experiments-simulation-ramp-comparison} illustrates this case. Reaching the goal required taking a detour down a ramp that could not be observed from the robot's starting position. As the virtual surface above the ramp was considered non-traversable, the robot never found a path to the goal. The robot only ever planned to the closest position within the space before the ramp and never observed or attempted traversal of the ramp.

When virtual surfaces were considered traversable the non-traversable negative obstacle could not be handled. The third column of costmaps in \Cref{fig:experiments-simulation-fatal-trench-comparison} illustrates this case. This is because the robot cannot possibly get any direct observations of the occluded vertical surface beyond the edge. The negative obstacle was always entirely virtual. Regardless of how close the robot got to the edge it could never directly observe the steepness of the negative obstacle below it. This resulted in the robot planning into the trench and falling.

When virtual surfaces were considered the best case slope for whatever lies below, both cases could be handled. The first column of costmaps in \Cref{fig:experiments-simulation-fatal-trench-comparison,fig:experiments-simulation-ramp-comparison} illustrate these cases. In \Cref{fig:experiments-simulation-fatal-trench-comparison} the robot approached the edge and identified the virtual surface as being representative of a non-traversable negative obstacle. In \Cref{fig:experiments-simulation-ramp-comparison} the robot planned onto the virtual surface above the ramp, moved toward the ramp, observed the ramp, then planned down the ramp and to the goal.

\Cref{table:experiments-simulation-values} contains a summary of simulations of 10 samples each of the scenarios described above. The success rate represents the fraction of samples where the robot behaved appropriately: for the trench scenario (\Cref{fig:experiments-simulation-fatal-trench-comparison}) success means not falling off the edge; for the ramp scenario (\Cref{fig:experiments-simulation-ramp-comparison}) success means reaching the goal. The duration is the amount of time taken by the robot to complete the test. A 60s timeout was set for the simulations so after 60s the sample was considered successful for the trench but failed for the ramp, hence the low std dev for the cases where this happened consistently. The low duration encountered when treating virtual surfaces as traversable in the trench case is due to the robot immediately driving off the edge. ROS bags recorded for these simulated test runs are publicly available for download.\footnote{\url{https://doi.org/10.25919/32q7-9e58}}.

\begin{figure}[t]
\begin{center}
\vspace{-5mm}
\includegraphics[width=0.6\columnwidth]{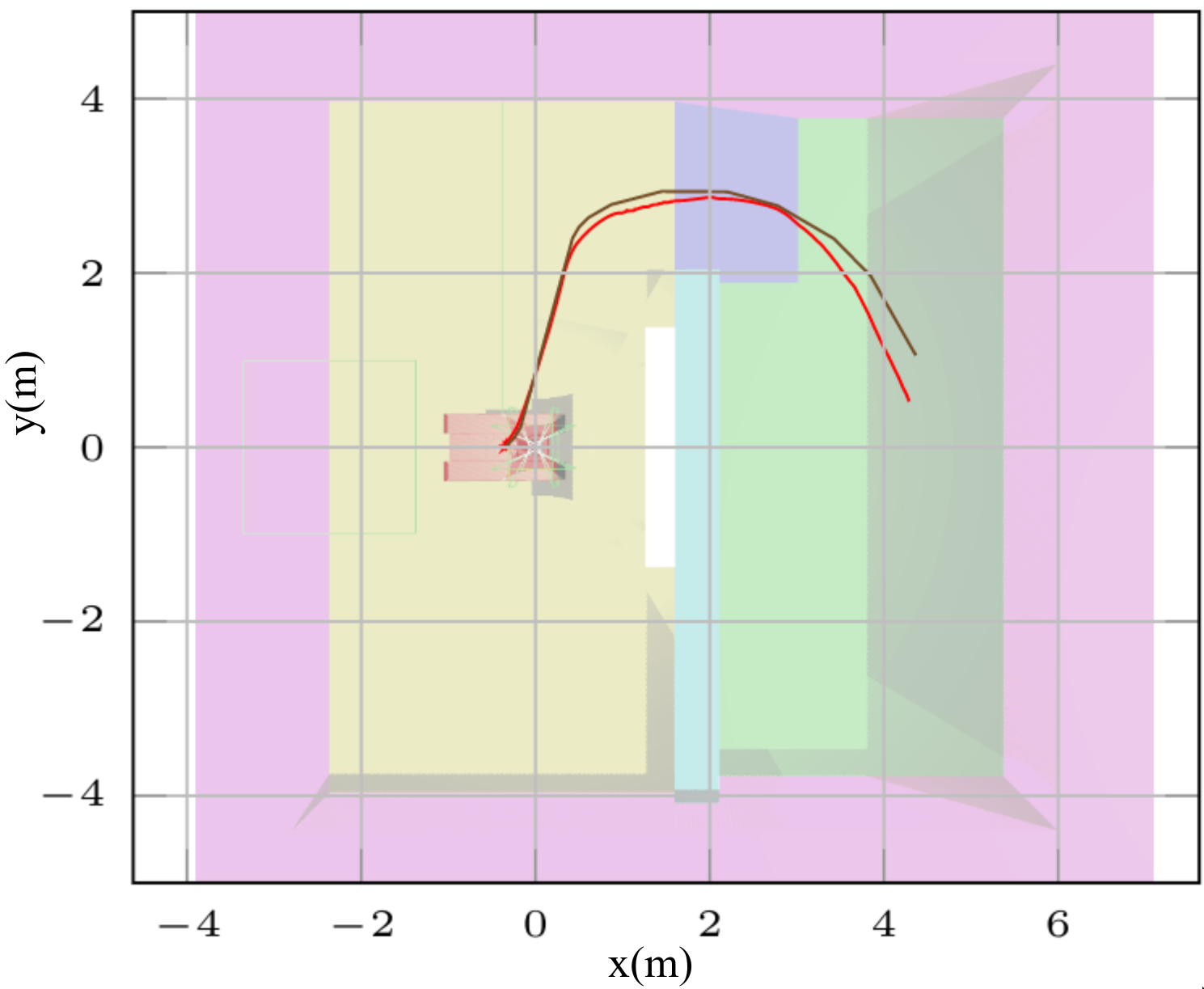}
\end{center}
\vspace{-5mm}
\caption{
An example of the first planned path (brown) and the actual robot trajectory (red) generated using virtual surfaces for the scenario in \Cref{fig:experiments-simulation-ramp-comparison}. The robot is continuously replanning so the divergence after $\sim$2\,m is due to a divergence in the plan not a failure in tracking.
}
\label{fig:experiments-simulation-ramp-path}
\end{figure}

\section{Discussions and limitations}
\label{sec:discussions}

\subsection{Virtual surfaces}
\label{sec:discussions-virtual-surfaces}

Limitations in the lidar driver make it impossible to accurately identify lidar samplings which do not yield a valid return within the sensor range. This is in part due to the ambiguity between returns which are beyond the sensor range and returns which lie within the sensor minimum range and in part due to the difficulty in determining the direction of the beam.

The lack of this information means that observations can only be made where valid returns are generated. Virtual surfaces can only be generated by such observations. Information may be missing from the map where no returns are generated. For example, if the robot approaches the edge of a platform over a large body of water, beams aimed at the water can get absorbed by the water due to the wavelength used for terrestrial Lidars, leading to no returns for estimating the water surface. The OHM map could generate virtual surfaces for these regions if information about the non-returning beams becomes available. This approach would adequately handle the scenario and the costmap could label the resulting virtual surfaces as non-traversable.


There is a significant delay (seconds) between the time when the robot is in a position capable of determining if a negative obstacle is fatal or not and the time when a costmap that includes that information is published. Currently, this delay is not considered, so the robot is at a significant risk of driving off the edge of a negative obstacle before it gets a chance to realise it is fatal. Firstly, this delay should be minimised in order to allow the robot to quickly determine fatality and move on. Secondly, this delay must be adequately handled in order to prevent driving onto virtual surfaces.

A downward slope is considered a negative obstacle when it is too steep to safely traverse. In order to determine if the slope is too steep, the robot must get close to the downward slope, it must get close to the region that it may not be able to safely traverse. This approach procedure is dangerous, especially for slopes that get steeper slowly such as driving off the top of a sphere. In many cases during testing, the robot ended up on a slope that it could not traverse without sliding downward toward the negative obstacle. Without handling the sliding correctly this inevitably leads to the robot falling down the slope. To better handle the approach, firstly the robot should move in line with the gradient of the slope so that it can back out safely without sliding and secondly it should stop when it is close enough to determine if it is too steep and wait for new observations to be made and processed.

\subsection{Planning}
\label{sec:discussions-planning}

\begin{figure}
\begin{center}
\includegraphics[width=0.95\columnwidth]{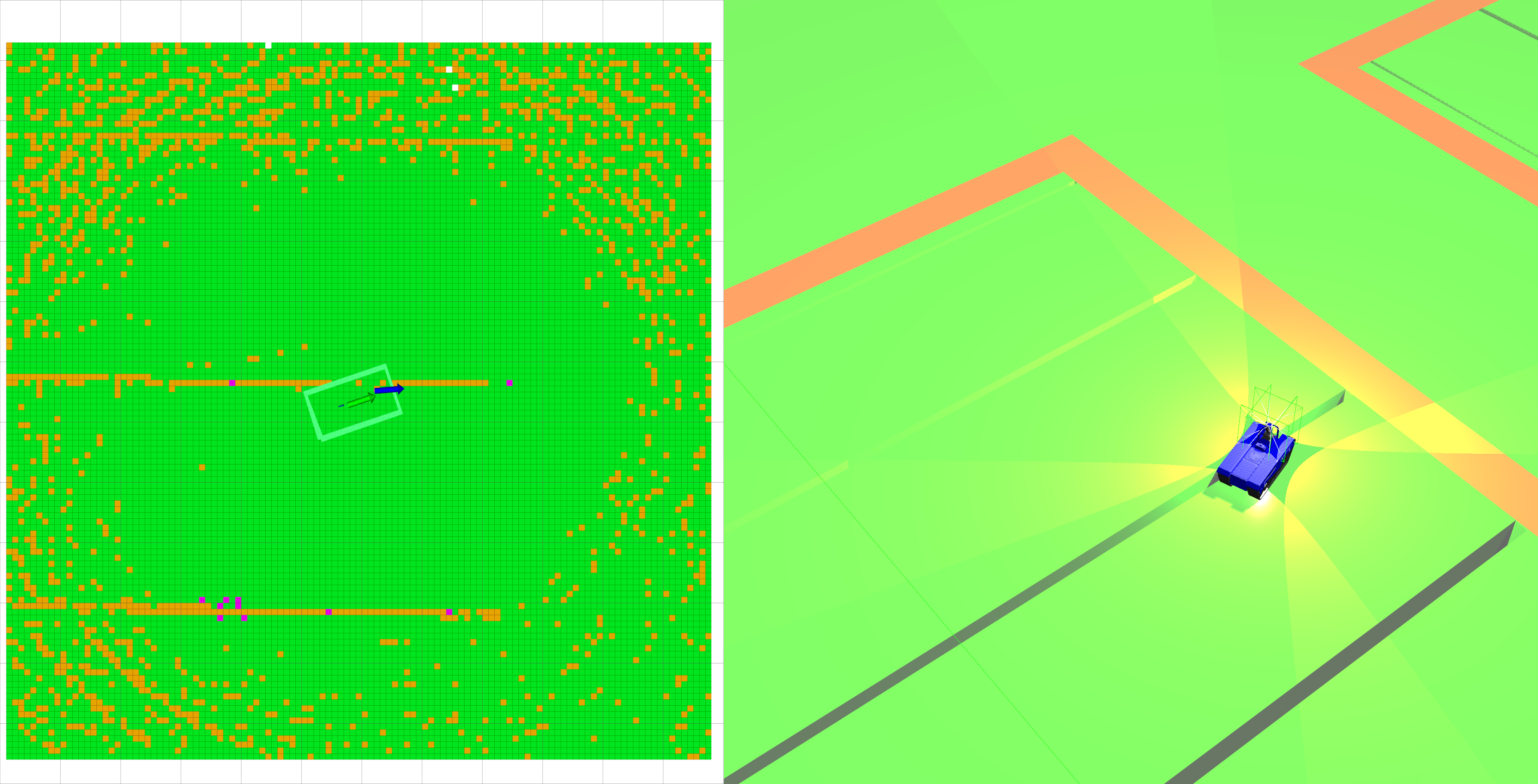}
\end{center}
\vspace{-5mm}
\caption{The end of a thin trench is inappropriately labelled as fatal due to being aligned to one of the sections.}
\label{fig:costmap-trench}
\vspace{-6mm}
\end{figure}

\begin{figure}
\begin{center}
\includegraphics[width=0.95\columnwidth]{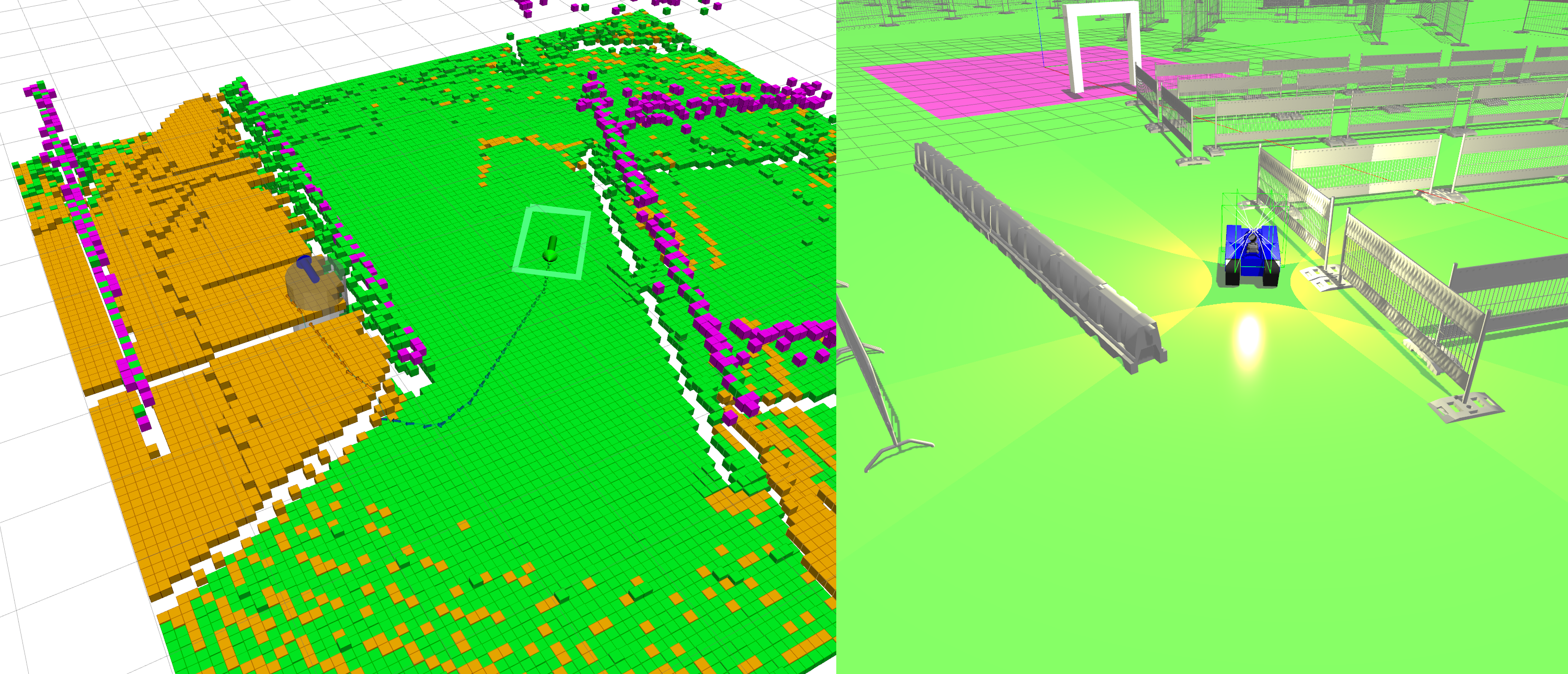}
\end{center}
\vspace{-5mm}
\caption{A virtual surface created by the shadow of a positive obstacle and a plan to move onto it.
}
\label{fig:costmap-positive-shadow}
\vspace{-5mm}
\end{figure}

The method used to label fatal obstacles in the costmap (see \Cref{sec:costmap}) is susceptible to incorrectly labelling the ends of thin but deep trenches as non-traversable where the robot can easily and safely traverse (see \Cref{fig:costmap-trench}). In addition, as only one method was used, it was tuned to handle both continuous slopes and discrete steps despite these two cases being represented significantly differently in the heightmap. To address this, one method should be used to detect non-traversable discrete steps and a different, independent method should be used to detect non-traversable continuous slopes. The method that was used in this work should not be used for either purpose.

The cost functions used in the costmap and in the planner both considered the robot to be a rigid rectangle capable of full control at all times. This lead to two problems. Firstly, the planner planned to traverse small obstacles that were taller than the vehicle's clearance but thinner than the gap between the vehicle's tracks. This often caused the vehicle to get stuck on top of such obstacles. Secondly, the planner assumes the vehicle can turn using full traction on both tracks at all times. When the vehicle is laterally traversing a slope (such that it is slightly rolled) the downhill track is supporting more weight than the uphill track. The downhill track has greater traction. Turning is not reliable when the left and right tracks have different amounts of traction. The planner should apply a penalty to making use of the uphill track when on a slope.

Virtual cells can be caused by the shadow of positive obstacles such as in \Cref{fig:costmap-positive-shadow}. In this case the virtual surface should be left as non-fatal to allow for planning around the corner. For this reason, all virtual cells are labelled as non-fatal in the costmap. Only the upper (higher elevation) end of slopes between cells are labelled as fatal. As a result, fatally steep virtual surfaces only lead to fatal costing where the virtual surface interfaces with real observations of a surface higher than them. This is a good description of the upper edge of a negative obstacle. Further work is required to address this in hybrid A* as the cost functions that were used may still give these virtual positive slopes a fatal cost.


Finally, the requirement to explore into unknown and possibly non-traversable areas means that the planner must be able to return partial paths which do not reach the goal it was given but get as close as possible. This poses significant risk near negative obstacles as it can mean driving the robot into the most unsafe possible non-fatal position. Improvements in candidate path selection when no complete path is possible could allow the robot to move into equivalent but significantly safer positions.

\section{Conclusions}
\label{sec:conclusions}

In this paper we have demonstrated a simple yet efficient technique for generating a best case surface prediction in regions of poor sensor observation. These predictions have been named virtual surfaces and can be effectively costed in path planning to generate valid paths which first approach, then avoid such poorly observed regions. We have also demonstrated how negative obstacles can be inferred by the presence of virtual surfaces and how these surfaces change on approach. Practical testing has built confidence that this approach is capable of handling a variety of difficult terrain.


\section*{Acknowledgements}
\label{sec:acknowledgement}

This work was funded by the U.S. Government under the DARPA Subterranean Challenge and by Australia's Commonwealth Scientific and Industrial Research Organisation (CSIRO). The authors would like to thank the entire CSIRO DATA61 DARPA SubT team and the Wildcat SLAM team for their support.





\balance
\bibliographystyle{IEEEtran}
\bibliography{references.bib}
\end{document}